\documentclass[conference]{IEEEtran}
\IEEEoverridecommandlockouts

\usepackage{cite}
\usepackage{amsmath,amssymb,amsfonts}
\usepackage{algorithmic}
\usepackage{graphicx}
\usepackage{textcomp}
\usepackage{xcolor}
\usepackage{cite}
\usepackage{hyperref}
\usepackage{url}
\usepackage{booktabs}
\usepackage{multirow}
\usepackage{array}
\usepackage{tabularx}
\usepackage{colortbl}
\usepackage[table]{xcolor}
\usepackage{amssymb}
\usepackage{tcolorbox}
\def\BibTeX{{\rm B\kern-.05em{\sc i\kern-.025em b}\kern-.08em
    T\kern-.1667em\lower.7ex\hbox{E}\kern-.125emX}}
\begin{document}

\title{Spectral-Aware Text-to-Time Series Generation with Billion-Scale Multimodal Meteorological Data
}

\author{\IEEEauthorblockN{Shijie Zhang}
\IEEEauthorblockA{\textit{School of Computer Science and Engineering, Northeastern University, Shenyang, China} \\
20235914@stu.neu.edu.cn}
}

\maketitle

\begin{abstract}
Text-to-time-series generation is particularly important in meteorology, where natural language offers intuitive control over complex, multi-scale atmospheric dynamics. Existing approaches are constrained by the lack of large-scale, physically grounded multimodal datasets and by architectures that overlook the spectral–temporal structure of weather signals. We address these challenges with a unified framework for text-guided meteorological time-series generation. First, we introduce MeteoCap-3B, a billion-scale weather dataset paired with expert-level captions constructed via a Multi-agent Collaborative Captioning (MACC) pipeline, yielding information-dense and physically consistent annotations. Building on this dataset, we propose MTransformer, a diffusion-based model that enables precise semantic control by mapping textual descriptions into multi-band spectral priors through a Spectral Prompt Generator, which guides generation via frequency-aware attention. Extensive experiments on real-world benchmarks demonstrate state-of-the-art generation quality, accurate cross-modal alignment, strong semantic controllability, and substantial gains in downstream forecasting under data-sparse and zero-shot settings. Additional results on general time-series benchmarks indicate that the proposed framework generalizes beyond meteorology.
\end{abstract}

\begin{IEEEkeywords}
Time Series Generation, Agentic AI
\end{IEEEkeywords}

\section{Introduction}
\label{sec:intro}
Generative modeling has achieved remarkable success in images~\cite{wang2025designdiffusion}, audio~\cite{jia2025audioeditor}, and text~\cite{zheng2025towards}, yet its potential for physical time series remains largely underexplored~\cite{cao2024survey,chen2025federated}. In meteorology, text-guided generation of weather time series is particularly valuable, as natural language provides an intuitive interface for controlling complex, multi-scale atmospheric dynamics. Such capability enables the simulation of rare extreme events~\cite{chen2023prompt,chen2024personalized,chen2024federated}, data augmentation for forecasting in observation-sparse regions~\cite{morid2023time}, and interactive climate analysis. Despite this promise, text-to-time-series generation poses challenges that fundamentally distinguish it from other modalities, arising from both data availability and model design.

A primary bottleneck lies in the absence of large-scale, high-quality datasets that pair time series with information-rich and physically meaningful textual descriptions. Unlike vision--language corpora such as LAION-5B~\cite{schuhmann2022laion}, existing time-series datasets (e.g., Time-MMD~\cite{liu2024time}, Time-IMM~\cite{chang2025time}) largely rely on rigid templates or heuristic rules. Consequently, their captions are often homogeneous and low in semantic density, describing surface-level values rather than underlying physical mechanisms. Such annotations rarely capture domain-specific reasoning, for example, distinguishing structured phenomena like diurnal cycles from stochastic fluctuations or explaining the physical causes of abrupt regime shifts. Without large-scale supervision that reflects diverse physical behaviors and causal semantics, models struggle to learn fine-grained cross-modal alignment between language and complex temporal dynamics.

Equally critical is the mismatch between existing generative architectures and the intrinsic structure of physical time series. Although diffusion models have achieved strong performance in other modalities~\cite{ho2020denoising}, their direct adaptation to time series often yields degraded results~\cite{yuan2024diffusion,lin2024diffusion,li2025bridge}. Meteorological signals are governed by superimposed frequency components spanning long-term trends, medium-term periodicities, and high-frequency variability~\cite{chen2023tempee,yi2024filternet,chen2023foundation,chenfedal,chen2022dynamic}. However, most text-conditioning strategies~\cite{ge2025t2s} operate purely in semantic space via cross-attention over raw text embeddings, lacking explicit inductive bias in the frequency domain. As a result, generated sequences may appear semantically plausible yet exhibit spectral distortions, failing to preserve periodicity, volatility, and long-range coherence. Bridging high-level linguistic instructions with low-level spectral--temporal dynamics therefore remains an open challenge.

To address these challenges, we propose a unified framework that advances both data construction and generative modeling for text-guided meteorological time series. First, we introduce \textbf{MeteoCap-3B}, a billion-scale weather dataset constructed via a Multi-agent Collaborative Captioning (MACC) pipeline. By coordinating LLM agents as perceptual analyzers, physical reasoners, and consistency critics, MACC produces physically grounded and information-dense captions that substantially surpass template-based annotations in diversity and domain fidelity. Second, we propose \textbf{MTransformer}, a diffusion generative model that explicitly incorporates spectral conditioning into the generation process. At its core, a \emph{Spectral Prompt Generator} translates natural language descriptions into multi-band spectral priors, which guide generation through a frequency-aware attention mechanism. This framework provides a new approach that jointly advances large-scale dataset construction and spectral-aware generative modeling for text-guided meteorological time series. Our contributions are:
\begin{figure*}[tbh]
    \centering
    \includegraphics[width=1\textwidth]{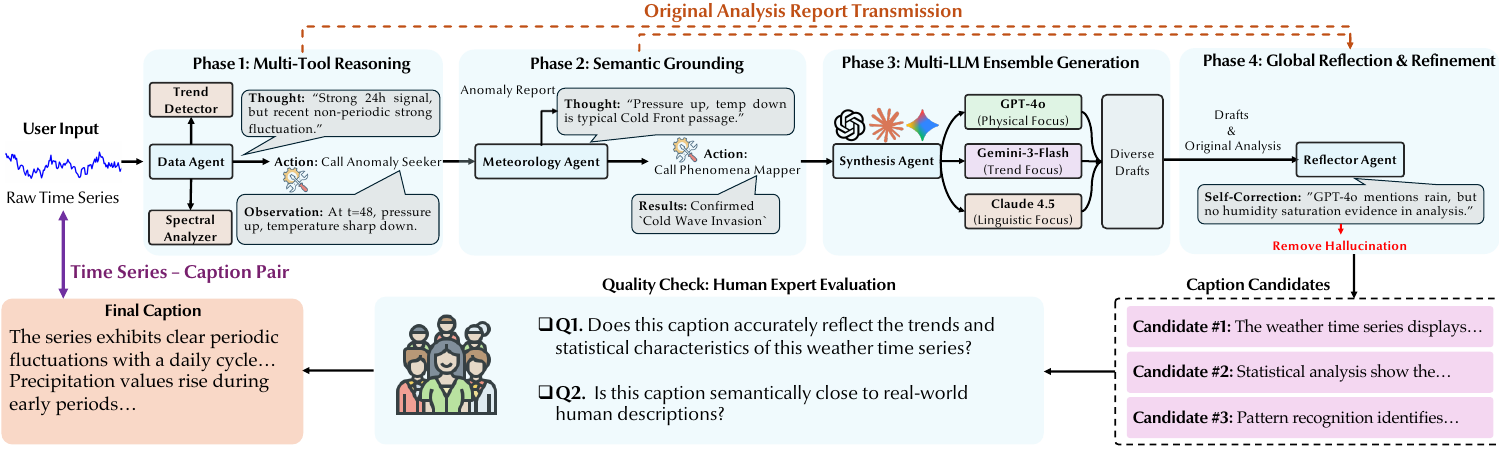}
    \caption{\small Overview of our Multi-agent Collaborative Captioning (MACC) pipeline for MeteoCap-3B construction. Note that we used GPT-4o, Claude Sonnet 3.5, and Gemini-3-Flash in Phase 3 and used DeepSeek V3~\cite{liu2024deepseek} in Phase 4. Human experts contains eight PhD in meteorology.}
    \vspace{-4pt}
    \label{fig:agent}
\end{figure*}
\begin{itemize}
\item We introduce the multi-agent-driven reasoning pipeline for automated weather time-series captioning, producing \textbf{MeteoCap-3B}, a large-scale, physically consistent multimodal dataset verified by human experts.
\item We propose \textbf{MTransformer}, a diffusion-based architecture that injects explicit spectral conditioning into the generation process, enabling precise natural-language control over multi-scale temporal dynamics.
\item Extensive experiments demonstrate state-of-the-art generation fidelity, accurate cross-modal alignment, and significant improvements in downstream forecasting under zero-shot and data-sparse settings.
\end{itemize}

\section{MeteoCap-3B: Agentic Captioning}
To bridge the modality gap between numerical weather observations and natural language, we construct \textbf{MeteoCap-3B}, a large-scale multimodal dataset comprising three billion meteorological observations paired with high-quality textual descriptions. The dataset is built based on our proposed \emph{Multi-agent Collaborative Captioning (MACC)} pipeline with human-in-the-loop refinement, ensuring both semantic richness and physical consistency, as shown in \textbf{Fig.~\ref{fig:agent}}.

\subsection{Multi-agent Collaborative Captioning}
The high-fidelity captioning for weather time series is formulated as a multi-stage collaborative process involving heterogenous LLM agents and specialized meteorological tools.

\noindent \textbf{Phase 1: Multi-Tool Reasoning.} 
The pipeline begins by extracting structured primitives from the raw sequence $\mathbf{x}_{1:T}$ under a ReAct-style workflow~\cite{yao2022react}. An agent iteratively interacts with a set of analytical tools $\mathcal{T}_{\text{perc}}=\{\text{Trend\_Det}, \text{Anomaly\_Seek}, \text{FFT\_Analyzer}\}$ to characterize temporal trends, abrupt changes, and dominant frequency components. For example, the agent computes the spectral energy distribution $\mathcal{E}_{\text{freq}}$ to identify periodicities and invokes the anomaly detector to locate abrupt shifts $\mathcal{A} = \{t | |\nabla x_t| > \delta\}$. The output of this phase is a structured perception vector $\mathbf{v}_{feat} = \text{Agent}(\mathbf{x}_{1:T}; \mathcal{T}_{perc})$ that summarizes the statistical and frequency-domain essence of the series

\begin{figure}[tbh]
\centering
\resizebox{\columnwidth}{!}{
\begin{tcolorbox}[colback=gray!5, colframe=black!75, title=\textbf{System Prompt: Phase 1}]
\small
\textbf{Role:} You are an expert meteorological data analyst.

\textbf{Goal:} Analyze the input time series $\mathbf{x}_{1:T}$ using available tools.

\textbf{Protocol:} 
\begin{enumerate}
\item Reason inside \texttt{<think>} tags before taking action.
\item Call tools inside \texttt{<tool>} tags.
\item Conclude with \texttt{<answer>} containing feature vector.
\end{enumerate}

\textbf{Input Series:} [288.15, 287.50, 285.20, ...] (Temperature K)

\textbf{Model Generation:}

\texttt{<think>}
The sequence shows a sudden drop. I need to check if this is a trend or a sudden anomaly. First, I will detect the overall trend.
\texttt{</think>}

\texttt{<tool>} call: Trend\_Detector(window=24) \texttt{</tool>}

\textit{[System: Tool Output: Trend=-0.85 (Strong Downward), p-value$<$0.01]}

\texttt{<think>}
The trend is significant. Now I need to check for periodicity to see if this is just a diurnal cycle.
\texttt{</think>}

\texttt{<tool>} call: FFT\_Analyzer(top\_k=3) \texttt{</tool>}
...
\end{tcolorbox}
}
\end{figure}

\noindent \textbf{Phase 2: Semantic Grounding.} The extracted primitives $\mathbf{v}_{\text{feat}}$ are then mapped to physically meaningful meteorological concepts. Equipped with a knowledge-retrieval tool $\mathcal{T}_{\text{kb}}$ encoding synoptic climatology rules, the agent performs symbolic reasoning over feature co-occurrences to infer high-level phenomena. This grounding step yields a semantic representation $\mathcal{K} = \text{Ground}(\mathbf{v}_{feat}, \mathcal{T}_{kb})$ containing interpretable tags such as \emph{frontogenesis}, \emph{diurnal thermal inversion}, or \emph{convective instability}, ensuring that subsequent captions reflect physically plausible interpretations rather than surface-level descriptions.

\begin{figure}[tbh]
\centering
\label{fig:prompt_phase2}
\resizebox{\columnwidth}{!}{
\begin{tcolorbox}[colback=white, colframe=blue!30!black, title=\textbf{System Prompt: Phase 2}]
\small
\textbf{Role:} You are a senior synoptic meteorologist.
\textbf{Task:} Interpret the provided numerical features using the retrieved climatological rules.
\textbf{Input:} 
\begin{enumerate}
\item Feature Vector: \texttt{{feature\_vector}} (e.g., trend=-0.8, fft\_peak=24h, anomaly\_idx=[48])
\item Retrieved Rules: \texttt{{retrieved\_rules}} (e.g., "Sharp temp drop + pressure rise to Cold Front")
\end{enumerate}

\textbf{Instruction:} 
- Do NOT generate the final caption yet.
- Output a list of \textbf{Physical Tags} and a brief \textbf{Reasoning Chain}.
- Verify if the anomaly aligns with the retrieved rules.

\textbf{Example Output:}
\texttt{<reasoning>}
The temperature trend is negative (-0.8), and there is a sharp anomaly at t=48. Combined with the retrieved rule, this strongly suggests a cold front passage disrupting the diurnal cycle.
\texttt{</reasoning>}
\texttt{<tags>} [Phenomenon: Cold\_Front], [Scale: Synoptic], [State: Transition\_Phase] \texttt{</tags>}
\end{tcolorbox}
}
\end{figure}

\noindent \textbf{Phase 3: Multi-LLM Ensemble Generation.} To capture diverse linguistic styles and focus points, we employ an \emph{Ensemble Generation Agent} that orchestrates multiple LLMs. Each model $\mathcal{M}_m \in \{\mathcal{M}_{GPT}, \mathcal{M}_{Clau}, \mathcal{M}_{Gemini}\}$ is prompted with the grounded knowledge $\mathcal{K}$ and a specific task-role (e.g., "Standard Reporting," "Trend Analysis," or "Casual Human-like Description"). This results in a candidate set of captions $\mathcal{C} = \{c_1, c_2, c_3\}$, where each $c_i = \text{Generate}(\mathcal{K}, \mathbf{x}_{1:T}; \mathcal{M}_i)$. This ensemble approach mitigates the stochastic bias of a single model and enriches the semantic diversity of the dataset.

\begin{figure}[tbh]
\centering
\label{fig:prompt_phase4}
\resizebox{\columnwidth}{!}{
\begin{tcolorbox}[colback=white, colframe=red!30!black, title=\textbf{System Prompt: Phase 4}]
\small
\textbf{Role:} You are a strict quality assurance auditor for meteorological reports.
\textbf{Input:} 
\begin{enumerate}
\item Original Data Facts: \texttt{{{feature\_vector}}} (Ground Truth)
\item Candidate Caption: \texttt{{{candidate\_caption}}}
\end{enumerate}

\textbf{Checklist:}
\begin{enumerate}
\item \textbf{Hallucination Check:} Does the caption mention events (e.g., "rain", "storm") not supported by the data features?
\item \textbf{Trend Consistency:} Does the described direction (up/down) match the numerical trend?
\item \textbf{Extrema Accuracy:} Are the mentioned peak times roughly correct?
\end{enumerate}

\textbf{Instruction:} 
- If errors are found, output \texttt{STATUS: REJECT} and provide specific \texttt{<feedback>}.
- If the caption is factual, output \texttt{STATUS: PASS}.

\textbf{Example Interaction:}

\textit{Input Caption:} "The region experienced heavy rainfall..."

\textit{Data Fact:} Precipitation variable is consistently 0.

\textit{Output:} 
\texttt{STATUS: REJECT}

\texttt{<feedback>} Hallucination detected: Caption claims rainfall, but data shows zero precipitation. Remove references to rain. \texttt{</feedback>}
\end{tcolorbox}
}
\end{figure}

\noindent \textbf{Phase 4: Global Reflection and Self-Correction.} The final phase involves a \emph{Reflector Agent} (powered by DeepSeek-V3) tasked with cross-consistency verification. The reflector takes $\mathcal{C}$ and the original data primitives $\mathbf{v}_{feat}$ as input to perform a ``fact-checking'' loop. It identifies hallucinations, such as an LLM claiming a ``heavy rain'' trend when the humidity sensor shows low saturation. The agent provides feedback $\mathcal{F}$ to the generation agents for iterative refinement: $c'_i = \text{Refine}(c_i, \mathcal{F})$. This phase ensures the internal validity of the captions before they are passed to the human-expert quality check stage, significantly reducing the human cognitive load.

\subsection{Human Expert Quality Check}
To ensure the fidelity and naturalness of the captions, we implement a rigorous HITL consensus mechanism. A panel of human meteorological experts evaluate $\mathcal{C}$ based on two orthogonal dimensions, including \textbf{Q1. Physical Alignment ($S_{pa}$):} Does the caption accurately reflect the trends, extrema, and statistical characteristics of $\mathbf{x}_{1:T}$? \textbf{Q2. Semantic Realism ($S_{sr}$):} Is the caption's linguistic structure consistent with professional human meteorological reporting? The final caption $c^*$ is selected or refined to maximize a joint utility function:
\begin{equation}
    c^* = \arg\max_{c \in \mathcal{C}} \left( \alpha \cdot \text{Sim}(\phi(\mathbf{x}_{1:T}), \psi(c)) + \beta \cdot \text{LLM-Eval}(c) \right),
\end{equation}
where $\phi(\cdot)$ is a pre-trained time-series encoder, $\psi(\cdot)$ is a text embedding Qwen3-Embedding-7B~\cite{yang2025qwen3}), and $\text{Sim}(\cdot, \cdot)$ denotes the cosine similarity in the joint embedding space.

\subsection{Complexity-based Categorization.} 
\begin{figure}[tbh]
    \centering
    \includegraphics[width=\columnwidth]{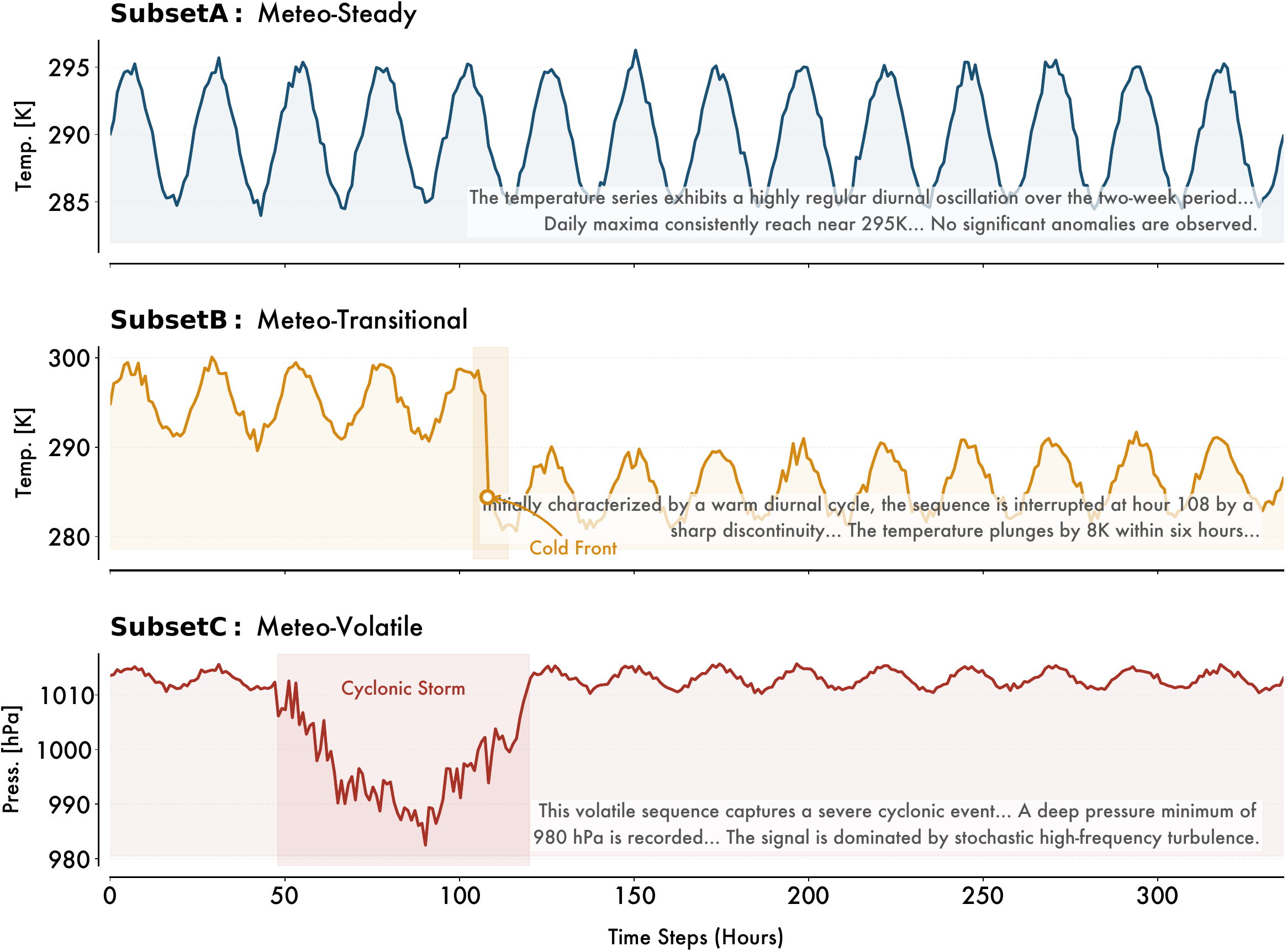}
    \caption{Representative samples across three subsets of MeteoCap-3B.}
    \label{fig:example}
    \vspace{-6pt}
\end{figure}

To facilitate robust training and evaluation, we categorize \textbf{MeteoCap-3B} into three subsets based on a \textit{Dynamical Complexity Index} ($\mathcal{D}$), computed as the weighted entropy of the signal's Lyapunov exponent and spectral volatility:
\begin{itemize}
    \item \textbf{Meteo-Steady ($\mathcal{D} < \delta_1$):} Sequences with high periodicity and low innovation, representing 60\% of the corpus.
    \item \textbf{Meteo-Transitional ($\delta_1 \le \mathcal{D} < \delta_2$):} Sequences featuring state transitions, representing 30\% of the corpus.
    \item \textbf{Meteo-Volatile ($\mathcal{D} \ge \delta_2$):} Sequences containing extreme weather events and stochastic turbulence, representing the remaining 10\%, as the most challenging benchmark.
\end{itemize}
Representative samples in MeteoCap-3B as shown in Fig.~\ref{fig:example}.

\subsection{Statistical Characteristics of MeteoCap-3B}
MeteoCap-3B is a large-scale multimodal meteorological corpus containing three billion observations organized into about one hundred million high-quality sequence–caption pairs. The data are curated from the Integrated Surface Database (ISD)~\cite{smith2011integrated}, covering thirty four years from 1991 to 2025 and more than thirty thousand stations worldwide. Each sequence $\mathbf{x}_{1:T}$ has length up to $T=512$, corresponding to one week of hourly measurements, and spans seven Köppen–Geiger climate zones. The dataset captures diverse temporal dynamics, including a Meteo-Steady subset with pronounced diurnal periodicity, where the average twenty four hour autocorrelation exceeds 0.85, and a Meteo-Volatile subset containing roughly 1.2 million extreme events such as rapid pressure drops above 24 hPa within twenty four hours. The accompanying textual modality forms a rich semantic layer, with over fifty thousand unique tokens and dense use of professional meteorological terminology. Each sequence is paired with multiple expert-written captions, and the final descriptions average forty five words with high linguistic complexity, reflected by a Flesch–Kincaid grade level of 10.5.

\begin{table}[tbh]
\centering
\caption{Statistics of MeteoCap-3B. Temp: temperature, Press: pressure, Humid: humidity, Wind: wind speed, Precip: precipitation.}
\label{tab:dataset_stats}
\resizebox{\columnwidth}{!}{
\begin{tabular}{llr}
\toprule
\textbf{Category} & \textbf{Metric} & \textbf{Value} \\
\midrule
\multirow{3}{*}{\textbf{Global Scale}}
& Total observation points & $3.15\times10^{9}+$ \\
& Number of stations & $31{,}400+$ \\
& Time span & 1991--2025 \\
\midrule
\multirow{3}{*}{\textbf{Series Modality}}
& Variables per step & 5 (Temp, Press, Humid, Wind, Precip) \\
& Sequence length ($T$) & 168 (hourly) / 512 (sub-hourly) \\
& Steady / Transient / Volatile & 60\% / 30\% / 10\% \\
\midrule
\multirow{3}{*}{\textbf{Text Modality}}
& Total unique captions & Approx. $105{,}000{,}000$ \\
& Average caption length & 45.2 words \\
& Vocabulary size & 52.4K \\
\midrule
\multirow{2}{*}{\textbf{Alignment}}
& Human QC pass rate & 94.2\% \\
& Avg. cosine similarity & 0.824 \\
\bottomrule
\end{tabular}
}
\vspace{-8pt}
\end{table}

\subsection{Spectral-aware Text-to-Time Series Generation}
\begin{figure}[tbh]
    \centering
    \includegraphics[width=1\columnwidth]{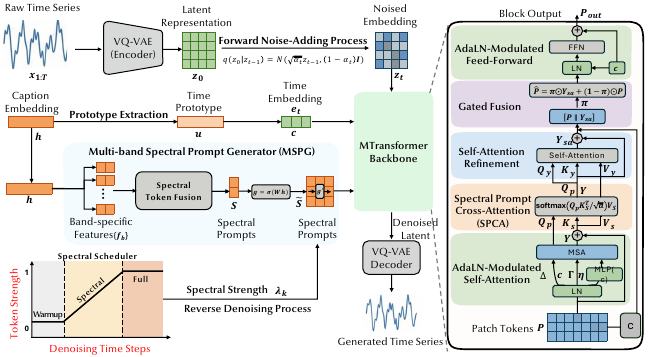}
    \caption{Our MTransformer for time series generation. \emph{Qwen-3-Embedding-7B}~\cite{yang2025qwen3} is used to achieve the caption embedding with 1024 dimensions.}
    \label{fig:model_arch}
\end{figure}

Directly adapting text-conditioned diffusion models to weather time series often neglects their intrinsic frequency structures, which are critical for physical consistency in meteorological data. To address this limitation, we propose a \emph{Spectral-aware Text-to-Time Series Diffusion} (\textbf{Fig.~\ref{fig:model_arch}}), which injects explicit multi-band spectral priors that are decoded from natural language captions into a Diffusion Transformer (DiT)~\cite{peebles2023scalable} operating in the latent space of a VQ-VAE~\cite{van2017neural}.

\noindent \textbf{Latent Diffusion Process.}
A raw sequence $\mathbf{x}_{1:T}$ is first encoded into a compact latent representation $\mathbf{z}_0 \in \mathbb{R}^{H\times W}$. Following DDPM~\cite{ho2020denoising}, the forward process gradually perturbs $\mathbf{z}_0$ with Gaussian noise:
$q(\mathbf{z}_t \mid \mathbf{z}_{t-1}) = \mathcal{N}(\sqrt{\alpha_t}\mathbf{z}_{t-1}, (1-\alpha_t)\mathbf{I})$.
The denoiser $\boldsymbol\epsilon_\theta$ is trained to recover $\mathbf{z}_0$, conditioned on both textual semantics and frequency-specific spectral prompts.

\noindent \textbf{Multi-band Spectral Prompt Generator (MSPG).}
To explicitly align textual descriptions with frequency-domain structures, we introduce the \emph{MSPG}. Given a caption embedding $\mathbf{h}$, MSPG projects $\mathbf{h}$ into $B$ frequency-specific subspaces:
\begin{equation}
\mathbf{f}_b = W_b \mathbf{h}, \quad b=\{1,\dots,B\},
\end{equation}
where $\{W_b\}$ are learnable projections corresponding to distinct frequency bands. Unlike fixed octave or predefined bands, these bands are \emph{learned end-to-end}, allowing the model to adaptively capture domain-specific temporal scales (e.g., diurnal, synoptic, or high-frequency variability) from data. For each band, a lightweight generator produces $M$ spectral tokens as
$\mathbf{S}_b = \operatorname{Gen}(\mathbf{f}_b) \in \mathbb{R}^{M\times d}$,
which are then concatenated and linearly fused:
$\mathbf{S} = \operatorname{Fuse}(\{\mathbf{S}_b\}_{b=1}^B) \in \mathbb{R}^{M\times d}.$ MSPG is trained jointly with the diffusion model via the standard denoising objective, without additional supervision, which encourages the generated spectral prompts to encode frequency priors that are most useful for latent denoising.

\noindent \textbf{Adaptive Spectral Gating.}
To allow selective activation of frequency bands based on textual semantics, we introduce an adaptive gating mechanism as follows:
\begin{equation}
\mathbf{g} = \sigma(W_g \mathbf{h}) \in \mathbb{R}^{M},
\end{equation}
where each element of $\mathbf{g}$ modulates the importance of a corresponding spectral token. The gated spectral prompts are given by $\tilde{\mathbf{S}} = \mathbf{S} \odot \mathbf{g}$. Semantically, this enables captions emphasizing, for example, ``volatile fluctuations`` or ``smooth trends`` to dynamically amplify high- or low-frequency components.

\begin{table*}[tbh]
 \centering
 \caption{\small Performance comparison on TSFragment-600K datasets and our proposed \textbf{\texttt{MeteoCap-3B}}. Note that for Meteo datasets, we evaluate on longer sequences (up to 336) to test long-range consistency. \textbf{Bold}: the best.}
 \begingroup
 \setlength{\tabcolsep}{3pt}
 \renewcommand{\arraystretch}{1.15}
 \definecolor{BestCell}{RGB}{225,245,230}
 \renewcommand{\bf}{\cellcolor{BestCell}\bfseries}
 \resizebox{1\textwidth}{!}{
 \begin{tabular}{c|ccccc|ccc|ccc|ccc|ccc|ccc|ccc}
 \toprule
 & & & \multicolumn{3}{c}{\textbf{MTransformer (Ours)}} & \multicolumn{3}{c}{\textbf{T2S~\cite{ge2025t2s} \textcolor{blue}{(IJCAI 2025)}}} & \multicolumn{3}{c}{\textbf{DiffusionTS~\cite{yuan2024diffusion} \textcolor{blue}{(ICLR 2024)}}} & \multicolumn{3}{c}{\textbf{TimeVAE~\cite{desai2021timevae} \textcolor{blue}{(arXiv 2025)}}} & \multicolumn{3}{c}{\textbf{GPT-4o-mini~\cite{hurst2024gpt} \textcolor{blue}{(2024)}}} & \multicolumn{3}{c}{\textbf{Llama3.1-8b~\cite{touvron2023llama} \textcolor{blue}{(2023)}}} & \multicolumn{3}{c}{\textbf{Gemini-3 \textcolor{blue}{(2025)}}} \\
 \cmidrule{4-24}
 \textbf{Group} & Datasets & Length & {WAPE $\downarrow$} & {MSE $\downarrow$} & {MRR@10 $\uparrow$} & {WAPE $\downarrow$} & {MSE $\downarrow$} & {MRR@10 $\uparrow$} & {WAPE $\downarrow$} & {MSE $\downarrow$} & {MRR@10 $\uparrow$} & {WAPE $\downarrow$} & {MSE $\downarrow$} & {MRR@10 $\uparrow$} & {WAPE $\downarrow$} & {MSE $\downarrow$} & {MRR@10 $\uparrow$} & {WAPE $\downarrow$} & {MSE $\downarrow$} & {MRR@10 $\uparrow$} & {WAPE $\downarrow$} & {MSE $\downarrow$} & {MRR@10 $\uparrow$} \\
 \midrule
 
 \multirow{18}[1]{*}{\rotatebox{90}{\textbf{TSFragment-600K~\cite{ge2025t2s}}}}
    & \multirow{3}[1]{*}{ETTh1} 
    & 24 & \bf 0.162 & 0.009  & 0.279 & {0.183} & \bf {0.008} & \bf {0.283} & 0.793 & 0.077 & 0.267 & 0.666 & 0.055 & 0.211 & 0.264 & 0.041 & 0.104 & 0.883 & 0.663 & 0.097 & 0.245 & 0.035 & 0.112 \\
    & & 48 & \bf 0.216 & \bf 0.014 & 0.300 & {0.234} & \bf{0.013} & 0.289 & 1.207 & 0.120 & {0.298} & 0.647 & 0.055 & 0.286 & 0.414 & 0.080 & 0.100 & 0.923 & 1.260 & 0.086 & 0.385 & 0.072 & 0.108 \\
    & & 96 & \bf 0.218 & 0.012 & \bf 0.296 & {0.229} & \bf {0.011} & {0.291} & 0.498 & 0.028 & 0.214 & 0.643 & 0.055 & 0.286 & 0.500 & 0.118 & 0.096 & 0.949 & 1.748 & 0.056 & 0.466 & 0.105 & 0.099 \\
 \cmidrule{2-24}
 
    & \multirow{3}[1]{*}{ETTm1}
    & 24 & 0.376 & 0.032 & \bf 0.287 & 0.426 & 0.033 & 0.286 & 0.604 & 0.040 & 0.251 & 0.666 & 0.048 & 0.219 & \bf 0.244 & \bf 0.031 & 0.101 & 1.134 & 0.798 & 0.099 & 0.238 & 0.030 & 0.108 \\
    & & 48 & \bf 0.368 & 0.054 & 0.280 & 0.53 & 0.053 & 0.283 & 1.119 & 0.100 & \bf 0.285 & 0.636 & \bf 0.051 & 0.217 & {0.453} & 0.112 & 0.097 & 1.074 & 1.496 & 0.079 & 0.425 & 0.101 & 0.103 \\
    & & 96 & \bf 0.389 & 0.036 & \bf 0.302 & 0.414 & 0.041 & 0.299 & 0.546 & \bf {0.031} & 0.293 & 0.664 & 0.057 & 0.208 & 0.706 & 0.395 & 0.091 & 1.079 & 1.761 & 0.057 & 0.655 & 0.310 & 0.095 \\
 \cmidrule{2-24}
 
    & \multirow{3}[1]{*}{Electricity} 
    & 24 & \bf 0.128 & 0.011 & \bf 0.282 & {0.135} & \bf {0.010} & {0.28} & 0.617 & 0.041 & 0.253 & 0.207 & 0.016 & 0.213 & 0.734 & 0.592 & 0.092 & 0.926 & 1.140 & 0.064 & 0.685 & 0.510 & 0.098 \\
    & & 48 & \bf 0.142 & \bf 0.012 & \bf 0.249 & 0.155 & 0.013 & {0.244} & 1.128 & 0.102 & 0.227 & 0.208 & 0.017 & 0.216 & 1.014 & 1.065 & 0.068 & 1.038 & 1.416 & 0.054 & 0.950 & 0.980 & 0.072 \\
    & & 96 & \bf 0.203 & 0.026 & \bf 0.314 & 0.238 & 0.031 & 0.318 & 0.545 & 0.032 & 0.247 & 0.213 & \bf {0.018} & 0.257 & 1.024 & 1.210 & 0.059 & 1.085 & 1.740 & 0.034 & 0.985 & 1.105 & 0.062 \\
 \cmidrule{2-24}
    & \multirow{3}[1]{*}{Exchange Rate} 
    & 24 & \bf 0.260 & \bf 0.031 & \bf 0.331 & {0.292} & {0.033} & {0.334} & 0.791 & 0.077 & 0.272 & 1.165 & 0.105 & 0.252 & 1.072 & 2.060 & 0.052 & 1.258 & 2.052 & 0.045 & 0.995 & 1.850 & 0.058 \\
    & & 48 & \bf 0.237 & \bf 0.029 & \bf 0.319 & 0.259 & {0.033} & {0.315} & 1.217 & 0.122 & 0.298 & 1.064 & 0.106 & 0.306 & 0.933 & 1.074 & 0.082 & 1.562 & 2.125 & 0.051 & 0.880 & 0.950 & 0.088 \\
    & & 96 & \bf 0.454 & \bf 0.043 & \bf 0.312 & {0.48} & {0.047} & {0.31} & 0.504 & 0.048 & 0.216 & 0.977 & 0.106 & 0.274 & 1.141 & 1.625 & 0.054 & 1.433 & 1.892 & 0.055 & 1.050 & 1.450 & 0.060 \\
 \cmidrule{2-24}
 
    & \multirow{3}[1]{*}{Air Quality} 
    & 24 & 0.672 & \bf 0.019 & \bf 0.313 & 0.884 & {0.02} & {0.304} & 0.806 & 0.078 & 0.265 & 2.303 & 0.022 & 0.302 & \bf {0.557} & 0.379 & 0.093 & 0.878 & 0.697 & 0.085 & 0.580 & 0.355 & 0.098 \\
    & & 48 & 1.245 & \bf 0.044 & \bf 0.312 & 1.295 & {0.044} & {0.297} & 1.439 & 0.120 & 0.221 & 1.648 & 0.023 & 0.271 & \bf {0.791} & 0.715 & 0.08 & 1.141 & 1.642 & 0.046 & 0.810 & 0.680 & 0.085 \\
    & & 96 & 1.348 & 0.043 & \bf 0.351 & 1.377 & 0.049 & {0.34} & \bf {0.508} & \bf 0.028 & 0.304 & 1.270 & {0.024} & 0.301 & 0.928 & 1.127 & 0.061 & 1.085 & 1.551 & 0.050 & 0.955 & 1.050 & 0.065 \\
 \cmidrule{2-24}
 
    & \multirow{3}[1]{*}{Traffic} 
    & 24 & \bf 0.331 & 0.006 & 0.213 & 0.353 & \bf {0.005} & 0.201 & 0.795 & 0.077 & 0.220 & 0.544 & 0.008 & {0.233} & 1.260 & 1.912 & 0.020 & 1.144 & 1.938 & 0.022 & 1.150 & 1.850 & 0.025 \\
    & & 48 & \bf 0.462 & \bf 0.007 & \bf 0.221 & {0.506} & {0.008} & {0.219} & 1.202 & 0.120 & 0.188 & 0.594 & 0.011 & 0.211 & 1.189 & 1.928 & 0.011 & 1.138 & 1.988 & 0.004 & 1.120 & 1.880 & 0.015 \\
    & & 96 & \bf 0.512 & \bf 0.009 & 0.259 & {0.543} & {0.01} & \bf {0.262} & 0.509 & 0.028 & 0.171 & 0.641 & 0.013 & 0.207 & 1.18 & 2.093 & 0.010 & 1.107 & 1.994 & 0.001 & 1.115 & 1.950 & 0.012 \\
 \midrule
 \midrule
 \multirow{9}[1]{*}{\rotatebox{90}{\textbf{MeteoCap-3B}}}
    & \multirow{3}[1]{*}{\textbf{\shortstack{Meteo\\Steady}}} 
    & 96 & \bf 0.092 & \bf 0.005 & \bf 0.345 & 0.108 & 0.007 & 0.330 & 0.450 & 0.025 & 0.240 & 0.135 & 0.012 & 0.310 & 0.350 & 0.065 & 0.125 & 0.850 & 0.620 & 0.110 & 0.320 & 0.055 & 0.135 \\
    & & 192 & \bf 0.115 & \bf 0.008 & \bf 0.328 & 0.142 & 0.010 & 0.305 & 0.580 & 0.038 & 0.210 & 0.168 & 0.015 & 0.285 & 0.420 & 0.095 & 0.110 & 0.980 & 0.850 & 0.095 & 0.405 & 0.088 & 0.118 \\
    & & 336 & \bf 0.138 & \bf 0.011 & \bf 0.312 & 0.175 & 0.014 & 0.288 & 0.650 & 0.045 & 0.190 & 0.210 & 0.020 & 0.250 & 0.550 & 0.145 & 0.095 & 1.150 & 1.100 & 0.080 & 0.510 & 0.130 & 0.105 \\
 \cmidrule{2-24}
 
    & \multirow{3}[1]{*}{\textbf{\shortstack{Meteo\\Trans.}}}
    & 96 & \bf 0.155 & \bf 0.012 & \bf 0.305 & 0.185 & 0.016 & 0.290 & 0.520 & 0.035 & 0.220 & 0.340 & 0.040 & 0.240 & 0.480 & 0.120 & 0.105 & 1.050 & 0.950 & 0.090 & 0.450 & 0.105 & 0.115 \\
    & & 192 & \bf 0.188 & \bf 0.018 & \bf 0.292 & 0.230 & 0.024 & 0.275 & 0.680 & 0.055 & 0.195 & 0.420 & 0.065 & 0.210 & 0.650 & 0.210 & 0.090 & 1.250 & 1.350 & 0.075 & 0.610 & 0.190 & 0.098 \\
    & & 336 & \bf 0.225 & \bf 0.025 & \bf 0.278 & 0.285 & 0.032 & 0.255 & 0.820 & 0.085 & 0.170 & 0.550 & 0.095 & 0.180 & 0.850 & 0.350 & 0.075 & 1.450 & 1.850 & 0.060 & 0.780 & 0.320 & 0.085 \\
 \cmidrule{2-24}
 
    & \multirow{3}[1]{*}{\textbf{\shortstack{Meteo\\Volatile}}} 
    & 96 & \bf 0.265 & \bf 0.028 & \bf 0.285 & 0.310 & 0.036 & 0.265 & 0.750 & 0.065 & 0.200 & 0.650 & 0.080 & 0.190 & 0.720 & 0.420 & 0.085 & 1.350 & 1.550 & 0.070 & 0.680 & 0.380 & 0.095 \\
    & & 192 & \bf 0.310 & \bf 0.038 & \bf 0.270 & 0.385 & 0.048 & 0.245 & 0.920 & 0.110 & 0.180 & 0.820 & 0.120 & 0.160 & 0.950 & 0.650 & 0.070 & 1.650 & 2.100 & 0.050 & 0.890 & 0.580 & 0.078 \\
    & & 336 & \bf 0.365 & \bf 0.052 & \bf 0.255 & 0.460 & 0.065 & 0.225 & 1.150 & 0.150 & 0.160 & 1.050 & 0.180 & 0.140 & 1.250 & 0.950 & 0.055 & 1.950 & 2.550 & 0.040 & 1.150 & 0.880 & 0.065 \\
 \midrule
 
 & \multicolumn{1}{c}{$1^{\text{st}}$ count} & - & \bf 23 & \bf 17 & \bf 22 & 0 & 5 & 2 & 1 & 2 & 1 & 0 & 2 & 0 & 3 & 1 & 0 & 0 & 0 & 0 & 0 & 0 & 0 \\
 \bottomrule
 \end{tabular}}
 \endgroup
 \vspace{-6pt}
\label{tab:main_results}
\end{table*}

\noindent \textbf{Denoising with Spectral-Prompt Cross-Attention.}
The denoiser is a Transformer. Each block first applies AdaLN-modulated self-attention to temporal patch tokens $\mathbf{P}$, followed by our proposed \emph{Spectral-Prompt Cross-Attention (SPCA)}:
\[
\bar{\mathbf{Y}} = \operatorname{Softmax}\!\left(\frac{(P W_Q)(\tilde{S} W_K)^\top}{\sqrt{d}}\right)(\tilde{S} W_V),
\]where $\mathbf{P}$ serves as Query and $\tilde{\mathbf{S}}$ as Key and Value. This design enforces frequency-aware refinement of the denoising trajectory, directly conditioning temporal updates on text-derived spectral priors. To prevent overly strong spectral constraints at early diffusion steps, we introduce a \emph{Spectral Scheduler} $\lambda_k$, which linearly increases from 0 to 1 over full training iterations. This weak-to-strong curriculum stabilizes optimization while progressively strengthening spectral guidance.

\section{Experiments and Results}
We design experiments to answer the following questions:
\begin{itemize}
    \item \textbf{RQ1:} How well does MTransformer generate high-quality time series compared with existing generative baselines across weather and general domains?
    \item \textbf{RQ2:} How does MTransformer scale with respect to model size, data scale, and sequence length?
    \item \textbf{RQ3:} Do the generated time series improve performance on downstream weather time series forecasting tasks?
    \item \textbf{RQ4:} Does MeteoCap-3B enable cross-modal alignment?
\end{itemize}

\noindent \textbf{Implementation.} During training, MeteoCap-3B is split into train, val, and test sets with a 7:2:1. The VQ-VAE is pretrained for 5k steps with 4 residual layers (hidden size 512, embedding size 128). The diffusion model is trained for 80k steps with spectral token length 6, text embedding size 1024, and text prototype size 128. The spectral training scheduler follows three phases: warmup ($0–20k$), spectral ($20k–40k$), and full ($40k–80k$). Our MTransformer uses a 12-layer Transformer with a hidden dimension of 768 and 12 attention heads, a feed-forward dimension of 3072, a temporal patch size of 16, and spectral prompts with 8 frequency bands and 16 tokens per band. Optimization is performed with AdamW using an initial learning rate of $1e^{-4}$ and a batch size of 256. The sampling steps are set to $1k$ unless otherwise specified. All implementations are conducted on 2 $\times$ NVIDIA A100 80GB.

\noindent \textbf{Baseline and Evaluation Metrics.} For time series generation, following~\cite{ge2025t2s}, we compare MTransformer with representative generative baselines, including T2S~\cite{ge2025t2s}, DiffusionTS~\cite{yuan2024diffusion}, TimeVAE~\cite{desai2021timevae}, and LLMs such as GPT-4o-mini~\cite{hurst2024gpt}, LLaMA-3.1~\cite{touvron2023llama}, and Gemini-3-Flash~\cite{ge2025t2s}. All methods are evaluated using three metrics: MSE, WAPE, and MRR@10. Experiments are conducted on two benchmarks: (i) TSFragment-600K~\cite{ge2025t2s}, which spans seven domains with 600K samples, and (ii) our \texttt{MeteoCap-3B}, comprising 3B high-quality samples organized into three subsets of increasing difficulty.

\subsection{RQ1: Evaluation on Text-to-Time Series Generation}
\noindent \textbf{Main Results.} \textbf{Table~\ref{tab:main_results}} summarizes the main results. MTransformer achieves the best performance in most settings, showing strong and stable generation quality. It consistently outperforms existing baselines on TSFragment-600K in WAPE and MRR@10 while maintaining competitive or best MSE, and exhibits a clearer advantage on \texttt{MeteoCap-3B} with lower errors across Steady, Transitional, and Volatile subsets, particularly under volatile conditions. These results demonstrate that MTransformer effectively handles long-range generation and challenging dynamics, outperforming both specialized time-series generators and general-purpose LLMs.
\begin{figure}[tbh]
    \centering
    \includegraphics[width=1\columnwidth]{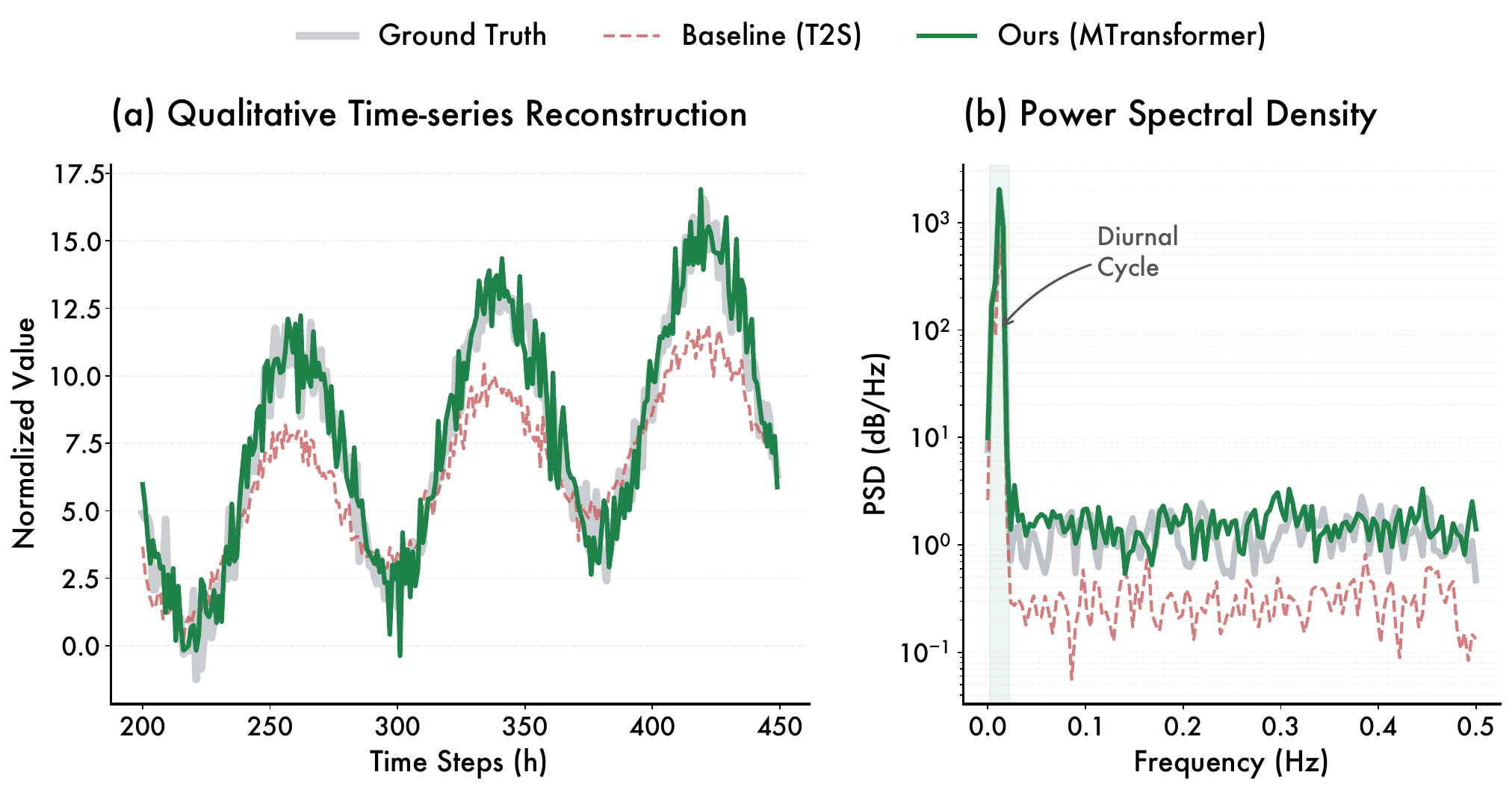}
    \caption{Qualitative analysis. \textbf{Left:} Qualitative time series reconstruction. \textbf{Right:} Power spectral density of real and generated time series.}
    \label{fig:qualitative}
\end{figure}

\begin{figure*}[tbh]
    \centering
    \includegraphics[width=.95\textwidth]{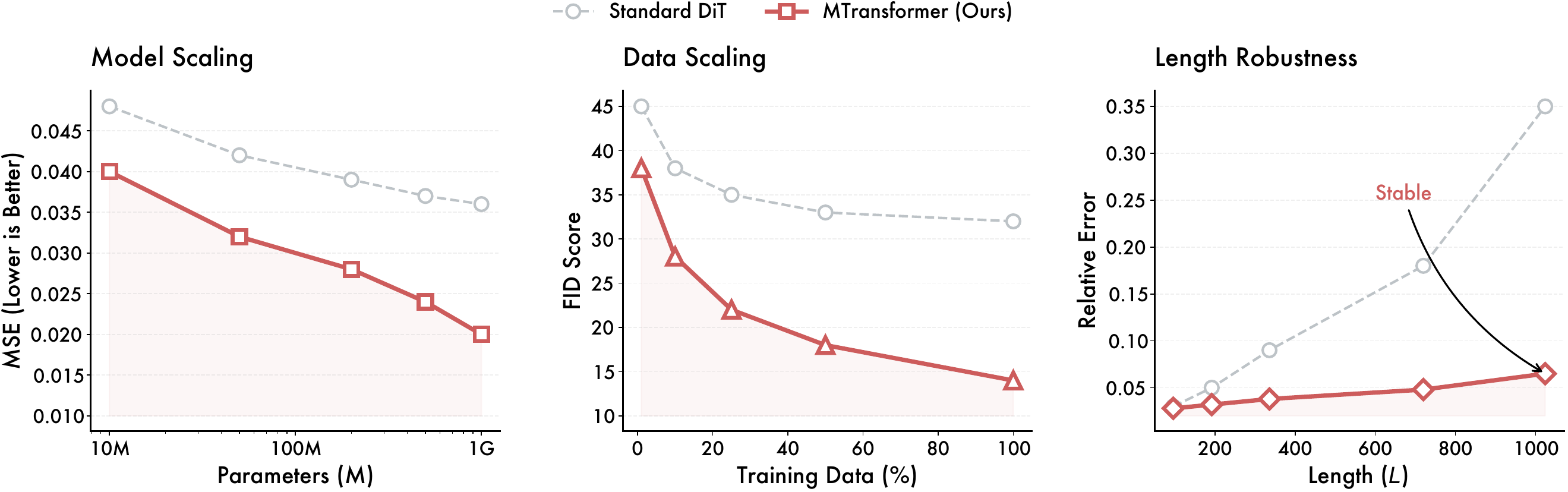}
    \caption{Scaling behavior of MTransformer with respect to model size (length of 96), data scale, and sequence length, evaluated on the Meteo-Volatile subset.}
    \label{fig:scaling}
    \vspace{-8pt}
\end{figure*}

\noindent \textbf{Qualitative Analysis.} 
\textbf{Fig.~\ref{fig:qualitative}} compares generation with the ground truth in time and frequency domains. MTransformer closely preserves trends and fluctuations in the time domain and accurately matches the ground-truth power spectral density across both low and high frequencies, while T2S smooths fine-grained dynamics and exhibits clear spectral distortion.

\subsection{RQ2: Evaluation on Model Scaling}
To further assess scalability, we analyze MTransformer along three dimensions using MSE as a unified metric: model size, data scale, and sequence length, as shown in Figure~\ref{fig:scaling}. For model scaling, MTransformer exhibits consistent performance gains as the parameter count increases from 10M to 1B, without clear saturation, in contrast to the standard DiT. For data scaling, MTransformer effectively exploits the large-scale MeteoCap-3B corpus, with a steeper performance improvement curve indicating higher data efficiency and lower MSE at comparable data budgets. Finally, when extending the sequence length from 96 up to 1024, standard models suffer from severe error accumulation, whereas MTransformer maintains stable and low MSE, demonstrating that the introduced spectral priors robustly anchor long-horizon generation. These demonstrate the superior scalability of our proposed MTransformer and provide insights into developing more scalable generative models for time series generation.

\subsection{RQ3: Evaluation on Downstream Weather Forecasting}
To assess the practical utility of the generated data, we conduct downstream forecasting experiments on the held-out ISD-Test dataset using TimesNet~\cite{wu2022timesnet} and a standard Transformer~\cite{vaswani2017attention} across four horizons $H \in \{96, 192, 336, 720\}$. The experiments evaluate both data augmentation and pre-training scenarios, using a 1:1 mix of synthetic and real data for augmentation. To prevent data leakage, all synthetic series are generated using only training-split metadata and temporal indices, with no access to test stations or future horizons.
\begin{table}[tbh]
    \vspace{-4pt}
\centering
\caption{Downstream forecasting performance (MSE $\downarrow$) on ISD-Test under the Few-shot (10\% Real Data) setting with two distinct backbones: a standard Transformer and TimesNet.}
\label{tab:augmentation}
\resizebox{\columnwidth}{!}{%
\begin{tabular}{l|cccc|cccc}
\toprule
\multirow{2}{*}{\textbf{Training Data Source}} & \multicolumn{4}{c|}{\textbf{Backbone: Transformer}} & \multicolumn{4}{c}{\textbf{Backbone: TimesNet}} \\
\cmidrule(lr){2-5} \cmidrule(lr){6-9}
 & \textbf{96} & \textbf{192} & \textbf{336} & \textbf{720} & \textbf{96} & \textbf{192} & \textbf{336} & \textbf{720} \\
\midrule
Real Only (10\% data) & 0.342 & 0.385 & 0.441 & 0.512 & 0.285 & 0.324 & 0.388 & 0.442 \\
\midrule
+ TimeVAE~\cite{desai2021timevae} & 0.335 & 0.372 & 0.430 & 0.498 & 0.278 & 0.315 & 0.375 & 0.430 \\
+ Diffusion-TS~\cite{yuan2024diffusion} & 0.318 & 0.355 & 0.412 & 0.475 & 0.262 & 0.301 & 0.358 & 0.412 \\
+ T2S~\cite{ge2025t2s} & 0.305 & 0.340 & 0.395 & 0.455 & 0.255 & 0.292 & 0.345 & 0.398 \\
 \textbf{+ MTransformer (Ours)} & \textbf{0.272} & \textbf{0.305} & \textbf{0.352} & \textbf{0.408} & \textbf{0.238} & \textbf{0.274} & \textbf{0.322} & \textbf{0.375} \\
\midrule
\textit{Reference: Real (100\%)} & \textit{0.255} & \textit{0.288} & \textit{0.335} & \textit{0.390} & \textit{0.215} & \textit{0.252} & \textit{0.304} & \textit{0.358} \\
\bottomrule
\end{tabular}%
}
\end{table}

\noindent \textbf{Augmentation for Data-Sparse Forecasting.} 
Table~\ref{tab:augmentation} reports forecasting MSE under a few-shot setting using only 10\% of real data. MTransformer consistently outperforms all competitive generative baselines across horizons and backbones. The advantage becomes more pronounced for long-term forecasting, indicating that the spectral-aware design effectively preserves low-frequency global trends that are often degraded in diffusion-based samples. MTransformer also surpasses T2S, demonstrating that explicit frequency constraints provide more reliable physical consistency than implicit text embeddings.

\noindent \textbf{Scalable Pre-training for Foundation Models.} 
We further evaluate MeteoCap-3B as a pre-training corpus. As shown in Table~\ref{tab:pretrain}, we pre-train an Transformer on the synthetic data and evaluate it under both Zero-Shot Transfer to unseen stations and Full Fine-Tuning. The multimodal pre-training strategy using aligned series–caption pairs achieves a 22.8\% reduction in MSE compared to training from scratch, demonstrating substantially improved cross-domain generalization. In contrast, unimodal pre-training yields only marginal gains, indicating that textual semantics provide effective regularization and that the synergy between text and time series is crucial for learning transferable representations.

\begin{table}[tbh]
\centering
\caption{Impact of Pre-training Modality on Downstream Performance under Zero-Shot Transfer and Full Fine-Tuning (FT).}
\vspace{-4pt}
\label{tab:pretrain}
\setlength{\tabcolsep}{5pt}
\resizebox{\columnwidth}{!}{
\begin{tabular}{l|cc|cc|cc}
\toprule
\multirow{2}{*}{\textbf{Pre-training}} & \multicolumn{2}{c|}{\textbf{Zero-Shot}} & \multicolumn{2}{c|}{\textbf{Full FT}} & \multicolumn{2}{c}{\textbf{Avg. Gain}} \\
 & \textbf{MSE} & \textbf{MAE} & \textbf{MSE} & \textbf{MAE} & \textbf{ZS} & \textbf{FT} \\
\midrule
\textbf{Scratch} (Random Init.) & 0.612 & 0.525 & 0.285 & 0.318 & - & - \\
\midrule
\textbf{Unimodal Pre-train} (Series Only) & 0.545 & 0.482 & 0.264 & 0.295 & +10.1\% & +7.3\% \\
\rowcolor{gray!15} \textbf{Multimodal Pre-train (Ours)} & \textbf{0.465} & \textbf{0.412} & \textbf{0.235} & \textbf{0.268} & \textbf{+22.8\%} & \textbf{+16.4\%} \\
\bottomrule
\end{tabular}
}
\vspace{-6pt}
\end{table}

\subsection{Framework Analysis}
\begin{figure}[tbh]
    \centering
    \includegraphics[width=1\columnwidth]{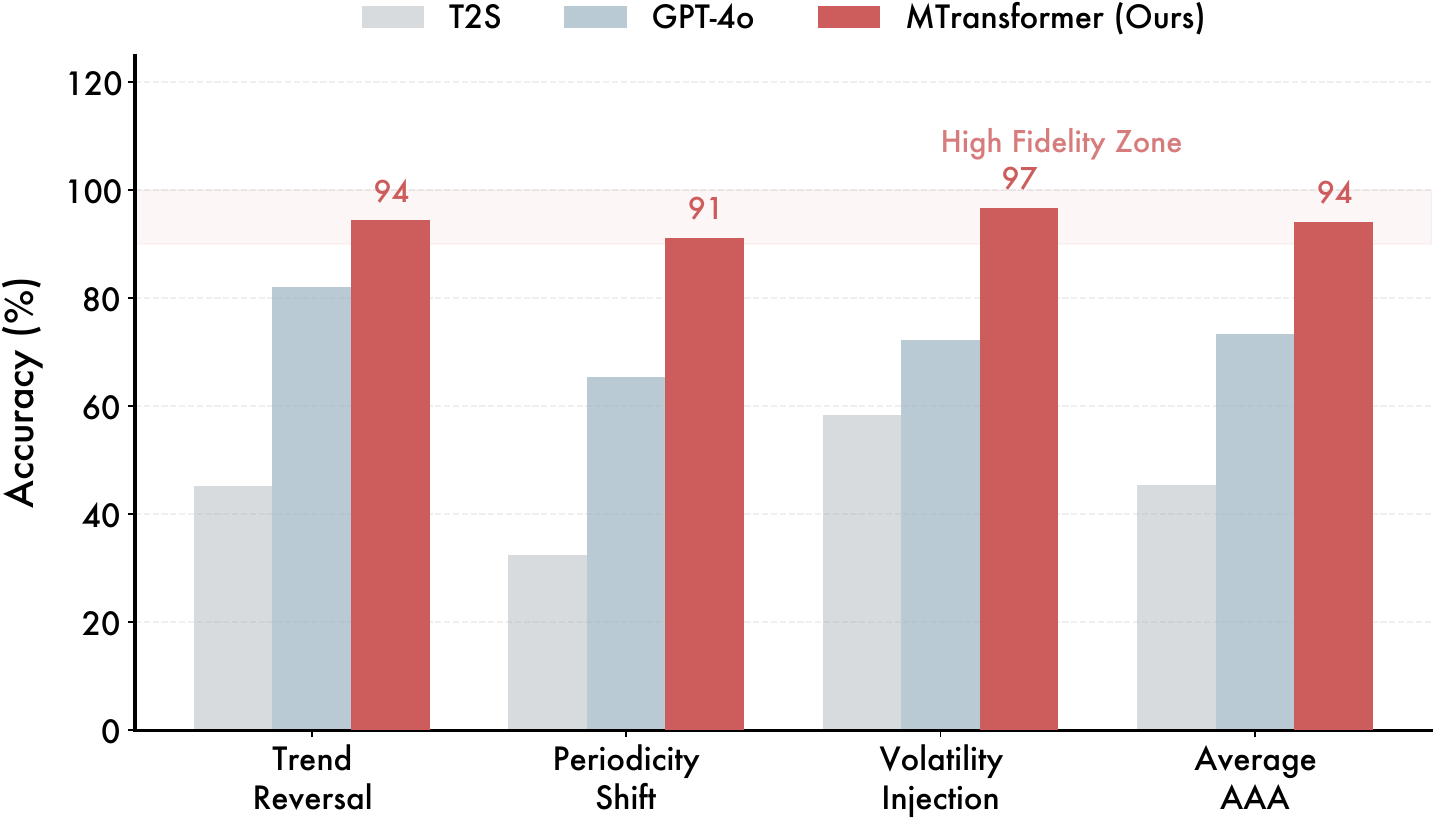}
    \caption{Controllability of MTransformer generated sequences.}
    \label{fig:controllability}
\end{figure}

\noindent \textbf{Semantic Controllability Evaluation.} \textbf{Fig.~\ref{fig:controllability}} reports counterfactual prompting results on 500 test cases, evaluating semantic controllability beyond distribution memorization. Using Attribute Alignment Accuracy (AAA) that verified via trend slopes, FFT peak shifts, and variance changes, MTransformer consistently achieves high alignment across all attributes, while T2S shows low sensitivity to textual control and GPT-4o struggles with precise long-horizon periodicity. These show that spectral prompting reliably translates frequency-level semantic instructions into corresponding temporal behaviors.

\noindent \textbf{Ablation Study.} Table~\ref{tab:ablation_sensitivity} (Part A) shows that each component is critical for performance on the Meteo-Volatile subset. Removing MSPG causes the largest degradation, which indicates that raw text embeddings fail to capture high-frequency variability without explicit spectral inductive bias. Replacing SPCA with standard cross-attention further increases MSE, highlighting its role in aligning temporal tokens with frequency-domain prompts. Disabling the scheduler leads to a smaller but consistent drop, suggesting that the weak-to-strong spectral curriculum stabilizes early-stage optimization.

\noindent \textbf{Hyperparameter Sensitivity.} 
Part B of Table~\ref{tab:ablation_sensitivity} examines the effect of spectral granularity. Increasing the number of frequency bands from $B=1$ to $B=8$ consistently improves performance, indicating better disentanglement of multi-scale weather patterns, while further increasing $B$ to 16 yields no additional benefit and likely introduces redundant high-frequency noise. Similarly, performance saturates at $M=16$ spectral tokens, with $M=32$ providing only marginal WAPE improvement at a higher computational cost. Based on this trade-off, we adopt $B=8$ and $M=16$ as the default setting.

\begin{table}[tbh]
\centering
\caption{Ablation study and hyperparameter sensitivity analysis on the \textbf{Meteo-Volatile} subset ($L=96$, MSE, WAPE report).}
\vspace{-4pt}
\label{tab:ablation_sensitivity}
\resizebox{0.815\columnwidth}{!}{%
\begin{tabular}{l|lcc}
\toprule
\multirow{6}{*}{\rotatebox{90}{\textbf{Part A: Ablation}}} & \textbf{Model Variants} & \textbf{MSE} $\downarrow$ & \textbf{WAPE} $\downarrow$ \\
\cmidrule{2-4}
 & \textbf{Full Model (Ours)} & \textbf{0.028} & \textbf{0.265} \\
 & \emph{w/o} Spectral Scheduler (Fixed $\lambda=1$) & 0.030 & 0.278 \\
 & \emph{w/o} SPAC (Standard Cross-Attn) & 0.033 & 0.292 \\
 & \emph{w/o} MSPG (Raw Text Embed.) & 0.038 & 0.325 \\
 & \emph{w/o} Text Cond. (Base DiT) & 0.045 & 0.368 \\
\midrule
\multirow{11}{*}{\rotatebox{90}{\textbf{Part B: Hyperparameter}}} & \textbf{Parameter Settings} & \textbf{MSE} $\downarrow$ & \textbf{WAPE} $\downarrow$ \\
\cmidrule{2-4}
 & \textit{Number of Frequency Bands ($B$)} & & \\
 & \quad $B=1$ (Global) & 0.035 & 0.312 \\
 & \quad $B=4$ & 0.029 & 0.275 \\
 & \quad $\mathbf{B=8}$ \textbf{(Default)} & \textbf{0.028} & \textbf{0.265} \\
 & \quad $B=16$ & 0.029 & 0.268 \\
\cmidrule{2-4}
 & \textit{Spectral Tokens per Band ($M$)} & & \\
 & \quad $M=4$ & 0.032 & 0.288 \\
 & \quad $M=8$ & 0.029 & 0.271 \\
 & \quad $\mathbf{M=16}$ \textbf{(Default)} & \textbf{0.028} & \textbf{0.265} \\
 & \quad $M=32$ & 0.028 & 0.264 \\
\bottomrule
\end{tabular}%
}
\vspace{-6pt}
\end{table}

\subsection{Validation of Captioning Pipeline} To validate the effectiveness of our captioning pipeline, we conduct a controlled human study on 400 randomly sampled instances. Captions generated by GPT-4o, Claude-4.5, and our MACC system are independently evaluated in a blind setting by eight PhD-level meteorology experts using three criteria: Hallucination Rate, measuring factual contradictions; Physical Consistency, rated on a 1–5 scale; and Informativeness, also rated on a 1–5 scale. MACC reduces hallucinations to 3.8\% while achieving near-perfect physical consistency and high informativeness, substantially outperforming all baselines. These results confirm that MACC delivers genuine gains in factual reliability and physical validity beyond simple API stacking.

\begin{table}[h]
\centering
\caption{Human expert evaluation of captioning quality. Values are reported as mean $\pm$ std.}
\label{tab:agent_eval}
\resizebox{1\columnwidth}{!}{
\begin{tabular}{l|ccc}
\toprule
\textbf{Method} & \textbf{Hallucination Rate} ($\downarrow$) & \textbf{Phys. Consistency} ($\uparrow$) & \textbf{Informativeness} ($\uparrow$) \\
\midrule
GPT-4o & 24.5\% & 3.12 $\pm$ 0.45 & 3.45 $\pm$ 0.38 \\
Claude-4.5 & 18.2\% & 3.35 $\pm$ 0.41 & 3.68 $\pm$ 0.42 \\
\rowcolor{gray!15} \textbf{MACC (Ours)} & \textbf{3.8\%} & \textbf{4.82 $\pm$ 0.15} & \textbf{4.65 $\pm$ 0.22} \\
\bottomrule
\end{tabular}
}
\end{table}
\begin{table}[tbp]
\centering
\caption{Bi-directional cross-modal retrieval performance on MeteoCap-3B test set. We compare against heuristic methods, trained baselines, and LLM-based adapters. \textbf{Meteo-Volatile} subset tests alignment under extreme conditions.}
\label{tab:retrieval_main}
 \definecolor{BestCell}{RGB}{225,245,230}
 \renewcommand{\bf}{\cellcolor{BestCell}\bfseries}
\resizebox{1\columnwidth}{!}{%
\begin{tabular}{l|ccc|ccc|ccc|ccc}
\toprule
\multirow{3}{*}{\textbf{Method}} & \multicolumn{6}{c|}{\textbf{Overall Test Set (All Subsets)}} & \multicolumn{6}{c}{\textbf{Meteo-Volatile Subset (Extreme Events)}} \\
\cmidrule(lr){2-7} \cmidrule(lr){8-13}
 & \multicolumn{3}{c|}{\textbf{Text-to-Series}} & \multicolumn{3}{c|}{\textbf{Series-to-Text}} & \multicolumn{3}{c|}{\textbf{Text-to-Series}} & \multicolumn{3}{c}{\textbf{Series-to-Text}} \\
 & \textbf{R@1} & \textbf{R@5} & \textbf{MRR} & \textbf{R@1} & \textbf{R@5} & \textbf{MRR} & \textbf{R@1} & \textbf{R@5} & \textbf{MRR} & \textbf{R@1} & \textbf{R@5} & \textbf{MRR} \\
\midrule
\multicolumn{13}{l}{\textit{Heuristic Baselines}} \\
Random Guess & 0.01 & 0.05 & 0.001 & 0.01 & 0.05 & 0.001 & 0.01 & 0.05 & 0.001 & 0.01 & 0.05 & 0.001 \\
Stat-Feature Matching & 8.5 & 22.1 & 0.145 & 8.2 & 21.5 & 0.142 & 4.2 & 12.8 & 0.085 & 3.8 & 11.5 & 0.080 \\
DTW-Matching & 12.4 & 26.8 & 0.185 & 11.5 & 25.2 & 0.178 & 6.8 & 15.5 & 0.110 & 6.2 & 14.2 & 0.105 \\
\midrule
\multicolumn{13}{l}{\textit{Neural Baselines}} \\
Time-MMD & 18.5 & 41.2 & 0.285 & 17.6 & 39.5 & 0.274 & 12.4 & 28.5 & 0.195 & 11.8 & 27.2 & 0.188 \\
TS-CLIP & 42.5 & 68.4 & 0.535 & 41.2 & 66.8 & 0.522 & 35.6 & 60.5 & 0.455 & 34.2 & 58.8 & 0.442 \\
LLM-Adapter & 48.2 & 73.5 & 0.585 & 47.5 & 71.8 & 0.572 & 40.5 & 65.2 & 0.495 & 39.2 & 63.8 & 0.482 \\
\midrule
{MTransformer (Ours)} & {\bf 58.2} & {\bf 81.4} & {\bf 0.675} & {\bf 56.9} & {\bf 79.8} & {\bf 0.662} & {\bf 52.5} & {\bf 76.2} & {\bf 0.615} & {\bf 50.8} & {\bf 74.5} & {\bf 0.598} \\
\emph{\quad w/o Spectral Prompt} & 45.8 & 72.1 & 0.565 & 44.2 & 70.5 & 0.552 & 38.5 & 64.2 & 0.485 & 37.2 & 62.5 & 0.468 \\
\bottomrule
\end{tabular}%
}
\end{table}
\subsection{RQ4: Cross-modal Alignment and Retrieval Analysis}
To verify whether MeteoCap-3B and our MTransformer effectively bridge the semantic gap between meteorological signals and natural language, we conduct bi-directional cross-modal retrieval experiments. This task evaluates the model`s ability to identify fine-grained correspondences between temporal dynamics and textual descriptions.

\noindent \textbf{Setup and Baselines.} 
We conduct retrieval on the held-out test split of MeteoCap-3B under two settings: \textbf{(1) Text-to-Series}, where a caption query retrieves time series via cosine similarity in the latent space, and \textbf{(2) Series-to-Text}, which performs the inverse retrieval. We report Recall@K ($K=1,5$) and MRR, and further evaluate on Meteo-Volatile to assess robustness under extreme conditions. We compare against three categories: \textbf{(1) Heuristics}, including \textit{Stat-Feature Matching} (Euclidean distance on statistical moments) and \textit{DTW-Matching}\cite{wang2022uncertainty}, which computes DTW distance between the query and a GPT-4o-generated pseudo-series; \textbf{(2) Trained Baselines}, including \textit{Time-MMD}\cite{liu2024time} (transfer) and \textit{TS-CLIP}\cite{chen2025ts}; and \textbf{(3) LLM Adapters}, represented by \textit{LLM-Adapter} built on frozen Llama-3\cite{touvron2023llama} with adapters~\cite{hu2023llm}.

\noindent \textbf{Results.} 
Table~\ref{tab:retrieval_main} reports the results. MTransformer achieves the best performance across all metrics, outperforming the strongest baseline, LLM-Adapter, by \textbf{+10.0\% in R@1}, with larger gains on \textbf{Meteo-Volatile}. TS-CLIP trained on MeteoScript-3B attains moderate performance while heuristic methods remain limited, highlighting the inadequacy of purely statistical or shape-based matching for meteorological semantics. The notable degradation without Spectral Prompts further underscores the importance of frequency-domain alignment.

\section{Conclusion}
In conclusion, we present a unified framework for text-guided meteorological time-series generation, comprising the \textbf{MeteoCap-3B} dataset constructed via a multi-agent collaborative pipeline and the MTransformer model. By integrating physically grounded multimodal supervision with explicit spectral conditioning, our approach achieves high-fidelity generation, accurate cross-modal alignment, and strong semantic controllability, while consistently improving downstream forecasting performance under data-sparse and zero-shot settings.


\bibliographystyle{IEEEtran}
\bibliography{IEEEabrv,reference}

\end{document}